\documentclass[journal]{IEEEtran}

\usepackage{cite}
\usepackage{graphicx}
\usepackage{amsmath}
\usepackage[caption=false,font=footnotesize]{subfig}
\usepackage{url}

\begin{document}

\title{Composing Music with Grammar Argumented Neural Networks and Note-Level Encoding}

\author{Zheng~Sun,
	Jiaqi~Liu,
	Zewang~Zhang,
	Jingwen~Chen,
	Zhao~Huo,
	Ching~Hua~Lee,
	and~Xiao~Zhang

\thanks{Z. Sun, J. Liu, Z. Zhang, J. Chen, and X. Zhang are with the Department of Physics, Sun Yat-sen University, Guangzhou, 510275 P. R. China (e-mail: \{sunzh6, liujq33, zhangzw3, chenjw93, zhangxiao\}@\{mail2, mail2, mail2, mail2, mail\}.sysu.edu.cn).}
\thanks{Z. Huo is with China University of Political Science and Law (e-mail: huozhao@cupl.edu.cn).}
\thanks{C. H. Lee is with the Institute of High Performance Computing (e-mail: calvin-lee@ihpc.a-star.edu.sg).}
\thanks{Manuscript received MM DD, YYYY; revised MM DD, YYYY.}}

\markboth{ }%
{Sun \MakeLowercase{\textit{et al.}}: Composing Music with Grammar Argumented Neural Networks and Note-Level Encoding}

\maketitle

\begin{abstract}
Creating aesthetically pleasing pieces of art, including music, has been a long-term goal for artificial intelligence research. Despite recent successes of long-short term memory (LSTM) recurrent neural networks (RNNs) in sequential learning, LSTM neural networks have not, by themselves, been able to generate natural-sounding music conforming to music theory. To transcend this inadequacy, we put forward a novel method for music composition that combines the LSTM with Grammars motivated by music theory. The main tenets of music theory are encoded as grammar argumented (GA) filters on the training data, such that the machine can be trained to generate music inheriting the naturalness of human-composed pieces from the original dataset while adhering to the rules of music theory. Unlike previous approaches, pitches and durations are encoded as one semantic entity, which we refer to as note-level encoding. This allows easy implementation of music theory grammars, as well as closer emulation of the thinking pattern of a musician. Although the GA rules are applied to the training data and never directly to the LSTM music generation, our machine still composes music that possess high incidences of diatonic scale notes, small pitch intervals and chords, in deference to music theory.
\end{abstract}


\IEEEpeerreviewmaketitle

\section{Introduction}
\IEEEPARstart{T}{he} creation of all forms of art\cite{van2016pixel,gatys2015neural,gregor2015draw,graves2013generating}, including music, has been a long-term pursuit of artificial intelligence (AI) research. Broadly speaking, music generation by AI is based on the principle that musical styles are in effect “complex systems of probability relationships”, as defined by the musicologist Leonard B. Meyer. In the early years, symbolic AI methods were popular and specific grammars describing a set of rules drive the composition\cite{rader1974method,Fern2013AI}. These methods were later much improved by evolutionary algorithms in various ways\cite{THYWISSEN1999GeNotator}, as embodied by the famous EMI project\cite{Cope1992Computer}. More recently, statistical models such as Markov chains and the Hidden Markov model (HMM) became popular in algorithmic composition\cite{allan2002harmonising}. Parallel to these developments was the rapid rise of neural network (NN) approaches, which have made remarkable progress in fields like signal and image recognition, as well as \cite{Silver2016Mastering} music composition. At present, the cutting-edge approaches to generative modeling of music are based on Recurrent Neural Networks (RNN)\cite{todd1989connectionist,MICHAEL1994Neural,boulanger2012modeling,Wermter2014Artificial} like the Long Short-Term Memory (LSTM)\cite{Franklin2006Recurrent,eck2008learning,jaques2016tuning} RNN.

While RNN and LSTM networks perform well in modeling sequential data, they suffer from a few significant shortcomings when applied to music composition. The music generated is often drab and dull without any discernible theme, consisting of notes that either sound either too repetitive or too random. It is thus desirable to have a machine that can learn to generate music adhering to the principles of music theory, although that is beyond the capabilities of ordinary neural networks or usual grammatical methods.

In this work, we hence improvise an LSTM with an original method known as the Grammar Argumented (GA) method, such that our model combines a neural network with grammars. We begin by training a LSTM neural network with a dataset from music composed by actual human musicians. In the training process, the machine learns the relationships within the sequential information as much as possible. Next we feed a short phrase of music to trigger the first phase of generation. Instead of adding the first phase of generated notes directly to the output, we evaluate these notes according to common music composition rules. Notes that go against music theory rules will be abandoned, and replaced by repredicted new notes that eventually conform to the rules. All amended results and their corresponding inputs will be then be added to training set. We then retrain our model with the updated training set and use the original generating method to do the second phase of (actual) generation. The abovementioned procedure is summarized in Fig. \ref{ga}. Another novel feature of our model is our note-level encoding method, which involves a new representation of notes by concatenating each note's duration and pitch as a single input vector. This combines the duration and pitch of each note as a single semantic entity, which is not only closer to how human composers think, but which also faciliates the direct application of music theory rules as grammars.

Our results indicate that our GA model possess markedly superior performance in music generation compared to its non-GA version, according to metrics based on music theory like the percentages of notes in the diatonic scale and chords, and pitch intervals within an octave. Indeed, our machine-created melodies sound pleasing and natural, as exemplified in our explicit example in Fig. \ref{result}. In all, our GA neural network with note-level encoding can learn basic music composition principles and produce natural and melodious music.

\section{Methods}
\subsection{Note-Level Encoding}
Although machine-learning methods have made significant progress in music composition, so far none has managed to closely simulate how human composers create music. In particular, human composers regard the pitch and duration of each note as attributes of a single entity, which in turn forms the building block of more complex musical motifs. 
By contrast, existing approaches either analyze pitches and note durations separately in separate neural networks\cite{Mozer1994, Franklin2006}, or represent music as quantized time series\cite{todd1989connectionist,boulanger2012modeling,Goel2014,eck2002finding,eck2008learning,Lyuq2015}. In this work, we shall attempt to more closely emulate human composers by combining the pitches and durations of musical notes into one entity, which we shall call as note-level encoding. Very importantly, this encoding allows the natural implementation of the rules of music theory as grammars, which act on notes and not merely fixed durations. This will be elaborated in Section 2.3. 

\begin{figure}[!t]
	\centering
	\includegraphics[width=3in]{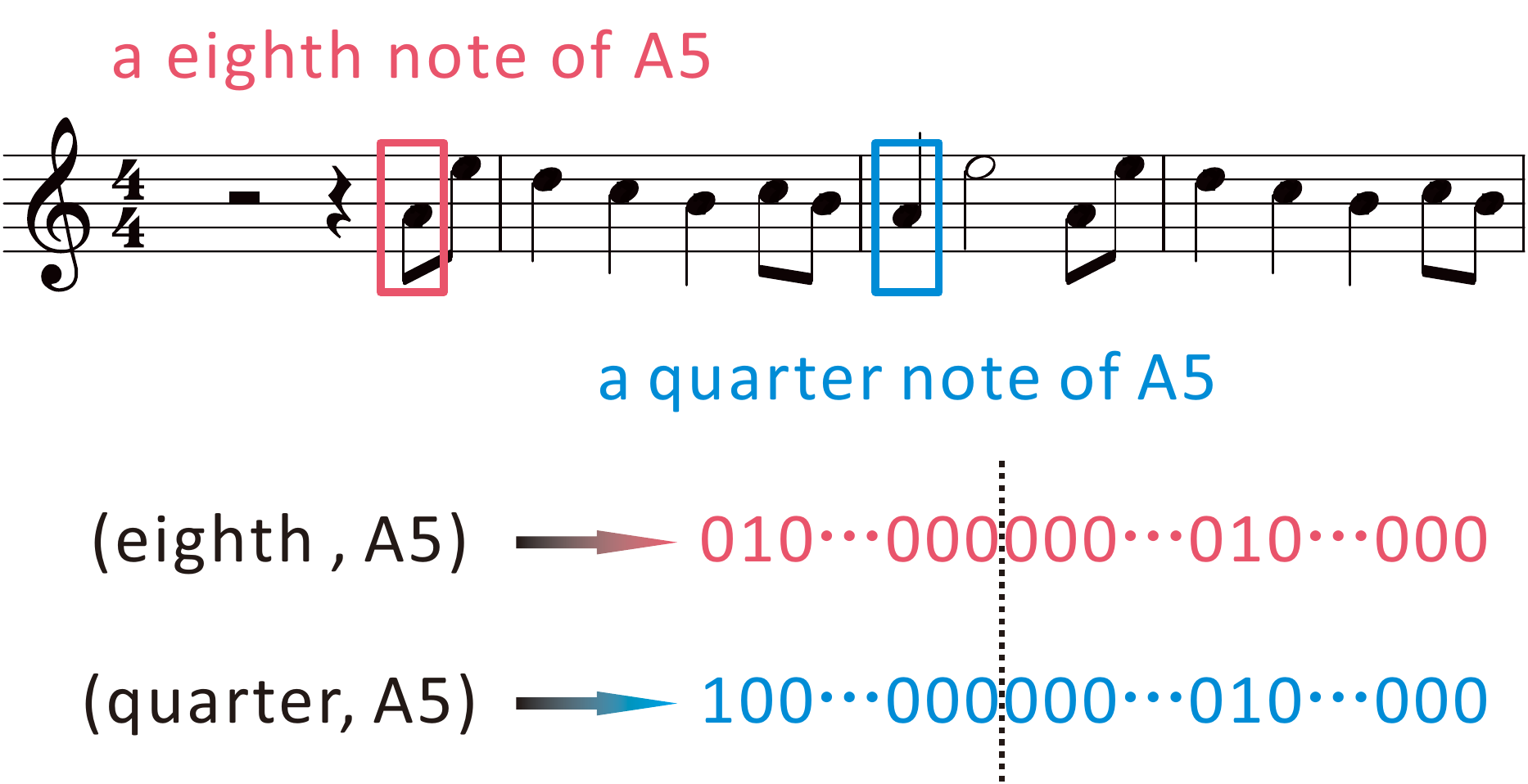}
	\caption{The duration and pitch of each note, as extracted from MIDI files, are encoded in a single (binary) one-hot vector, This is illustrated by the quarter and eighth notes above, both of A5 pitch.}
	\label{encoding}
\end{figure}

Our training data is derived from the MIDI sequences of 106 piano pieces by contemporary musicians like Joe Hisaishi, Yiruma, Yoko Kanno and Shi Jin. For consistency, we transpose all pieces to start with C major/A minor, only include pieces with 4/4 time signature, and retain only the melody such that the resultant music is monophonic. This entails omitting music accompaniments, grace notes and intensity changes. In particular, only the highest note, which typically carries the melody, is retained when simultaneous notes occur. This leaves us with a sequence of "Note On Events" and "Note Off Events", which can then be directly encoded as a sequence of one-hot vectors containing duration and pitch information, like Fig. \ref{encoding}. Each one-hot vector consist of a 59-bit segment representing pitch semitones from A0 to C8, concatenated with a 30-bit segment representing durations from a semiquaver to a breve. Indeed, by including both pitch and duration within a single vector, our note-level encoding method enables the machine to "learn" music composition by regarding the notes as fundamental building blocks, just like with human composers. 

\subsection{Long Short-Term Memory Neural Networks}
Recurrent neural networks (RNN) are widely used in sequential learning. They "remember" information from previous time steps as each of their hidden layers receives input from the previous layer, as well as input from itself one time step ago. However, simple RNNs are inadequate for music composition as they do not perform well with long-term dependency due vanishing gradients\cite{Hochreiter1997}. This long-term dependency is necessary for understanding musical motifs which often last beyond several time steps. Our solution is to employ a more advanced type of RNN known as a long short-term memory (LSTM) neural network, which also possess a memory cell with potentially longer-term storage of data controlled by various gates.

\begin{figure}[!t]
	\centering
	\subfloat[ ]{\includegraphics[width=1.25in]{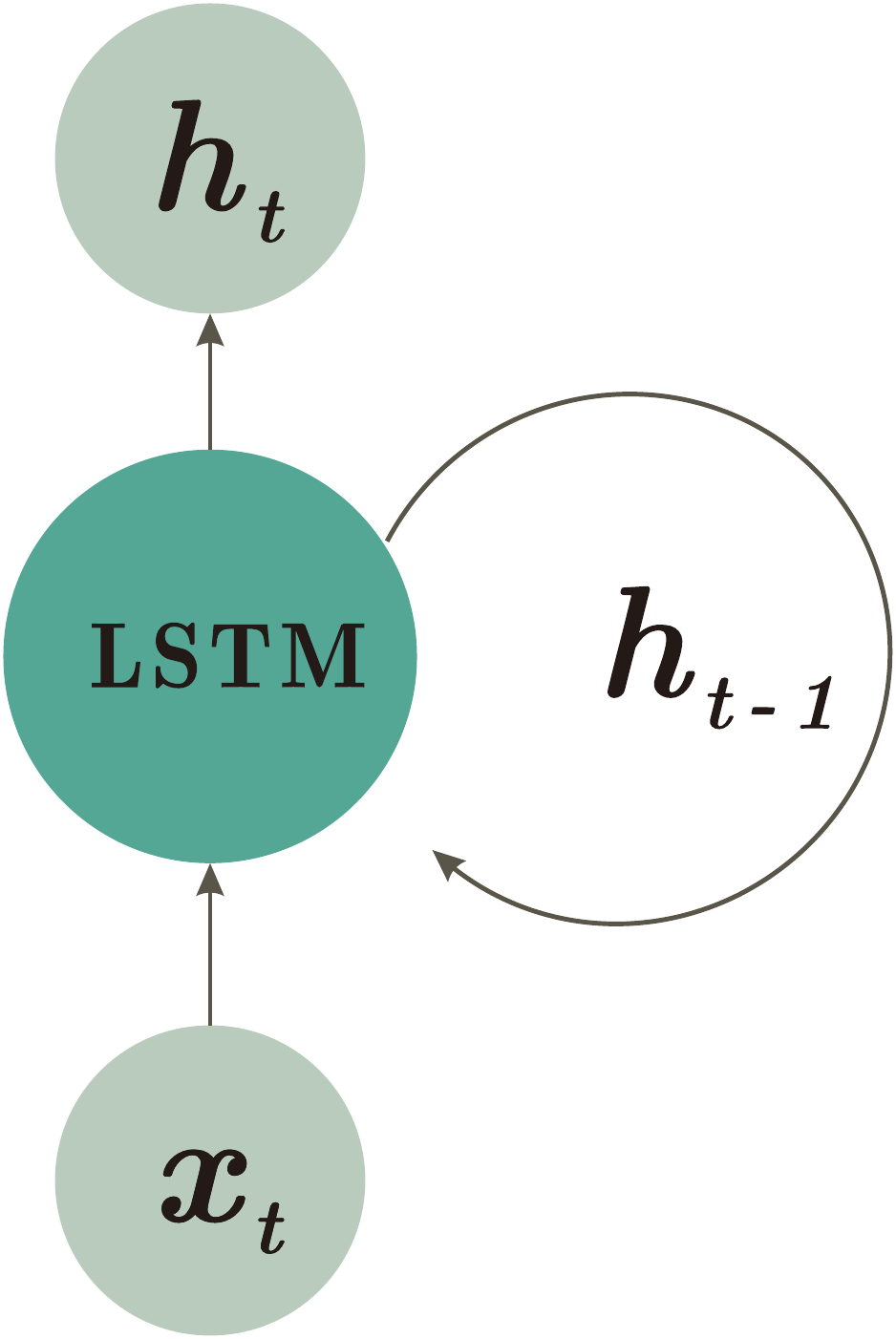}%
		\label{lstma}}
	\hfil
	\subfloat[ ]{\includegraphics[width=2.5in]{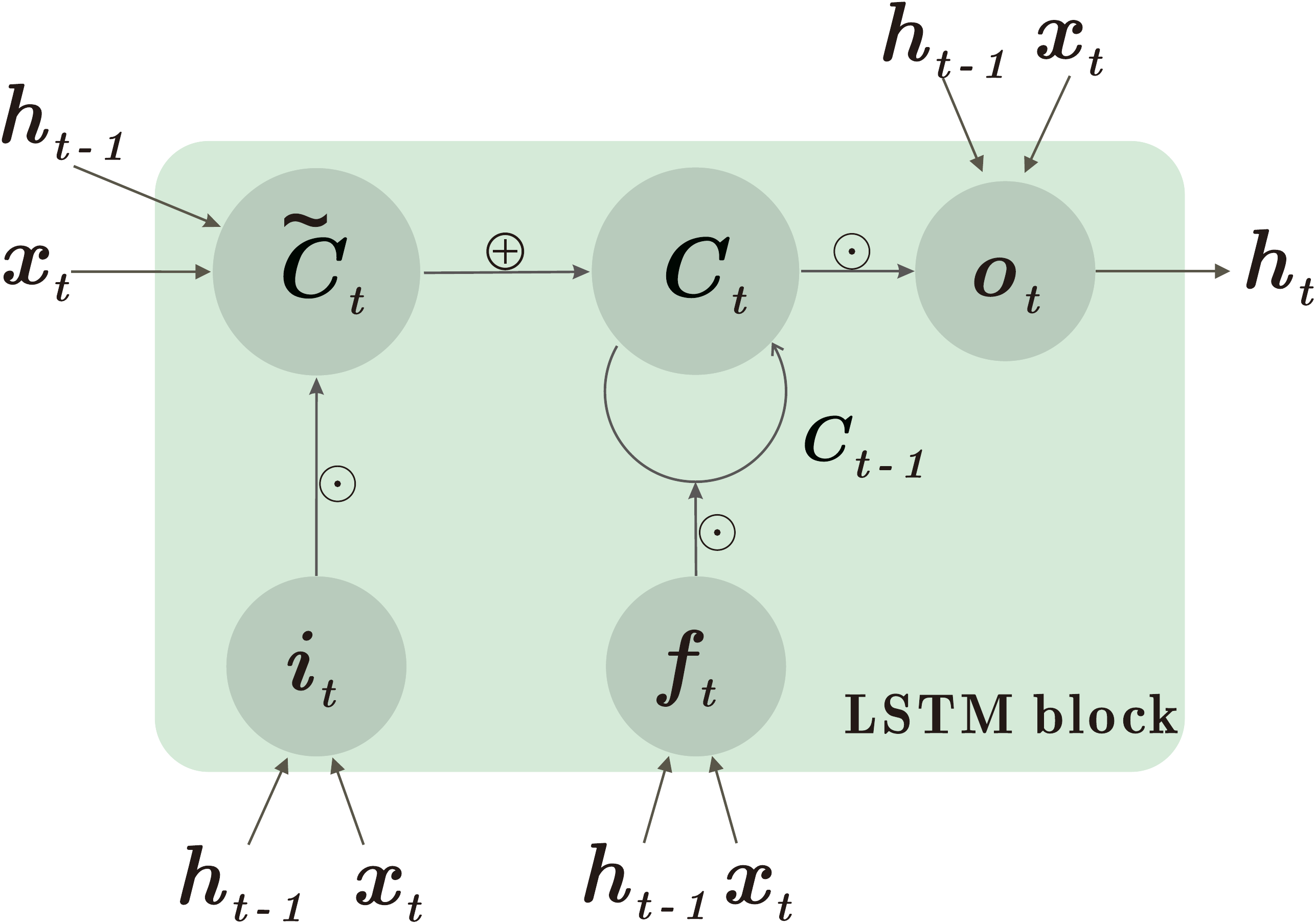}%
		\label{lstmb}}
	\caption{The structure of the LSTM module. With the information from input and hidden states $x_{t}$ and $h_{t-1}$, an LSTM layer outputs a hidden state $h_{t}$ that conveys temporal information. $\odot$ denotes element-wise multiplication and $\oplus$ denotes element-wise addition. The upper diagram illustrates the logical dependencies of the input and outputs, and the lower diagram schematically illustrates the how an LSTM layer compute the current hidden state $h_{t}$.}
	\label{lstm}
\end{figure}

An LSTM module contains a memory cell state $C_{t}$ in addition to its hidden state $h_{t}$, as in Fig. \ref{lstm}. Unlike the hidden state, $C_{t}$ is linearly related to its past values, and can thus store information for an arbitrary duration until they are "erased" by the forget gate. At each time step, the values of the input $x_{t}$, previous memory cell state $C_{t-1}$ and previous hidden state $h_{t-1}$ together determine the new memory cell state $C_{t}$ and new hidden state $h_{t}$. This achieved with the input gate $i_{t}$, output gate $o_{t}$ and forget gate $f_{t}$ defined by 
\begin{equation}
i_{t} = \sigma(W_ix_{t} + U_ih_{t-1} + b_i)
\end{equation}
\begin{equation}
o_{t} = \sigma(W_ox_{t} + U_oh_{t-1} + b_o)
\end{equation}
\begin{equation}
f_{t} = \sigma(W_fx_{t} + U_fh_{t-1} + b_f)
\end{equation}
where $b_i,b_o$ and $b_f$ are the corresponding vectors of biases,  $W_i, W_o$ and $W_f$ are the corresponding weight matrices for the input vectors and $U_i,U_o$ and $U_f$ are the corresponding weights connecting the previous hidden state vectors. The element-wise sigmoid function $\sigma(x) = (1 + e^{-x})^{-1}$ realizes the filtering role of the gates with its output value increasing from $0$ (block) to $1$ (pass) as the input ranges from $-\infty$ to $+\infty$. At each time step, the memory cell state is updated according to 
\begin{equation}
C_{t} = i_{t} \odot \widetilde{C}_{t} + f_{t} \odot C_{t-1}
\end{equation}

The forget gate $f_{t}$ controls how much information is "forgotten" i.e. not passed on: if $f_{t}$ is zero, all previous information $C_{t-1}$ in the memory cell is forgotten. The input gate $i_{t}$ controls the amount of "new" input to the memory cell from the activated current state memory $\tilde C_{t}$ defined by 
\begin{equation}
\widetilde{C}_{t} = \tanh(W_cx_{t} + U_ch_{t-1} + b_c)
\end{equation}
which depends on the current input and most recent hidden state data. $W_c$, $U_c$ and $b_c$ are the associated weight matrices and biases respectively.

Finally, the hidden state $h_{t}$ of the LSTM is updated according to the activated current state of memory cell under the control of output gate:
\begin{equation}
h_{t} = o_{t} \odot \tanh(C_{t})
\end{equation}

As a differentiable function approximator, the (weights and biases of the) LSTM are typically trained with gradient descent\cite{Hochreiter1997}, with gradient calculated via Back-propagation Through Time (BPTT)\cite{graves2005framewise}. The training details of our LSTM will be discussed in Section \ref{sec:ex}.

\subsection{Grammar Argumented Method}\label{sec:ga}
One problem plaguing neural network approaches to music composition is that the music generated largly do not conform to basic principles of music theory. For instance, they often have too many overtones (excessive chromaticity), overly large pitch intervals, and unharmonious melodies.

\begin{figure*}[!t]
	\centering
	\includegraphics[width=5.5in]{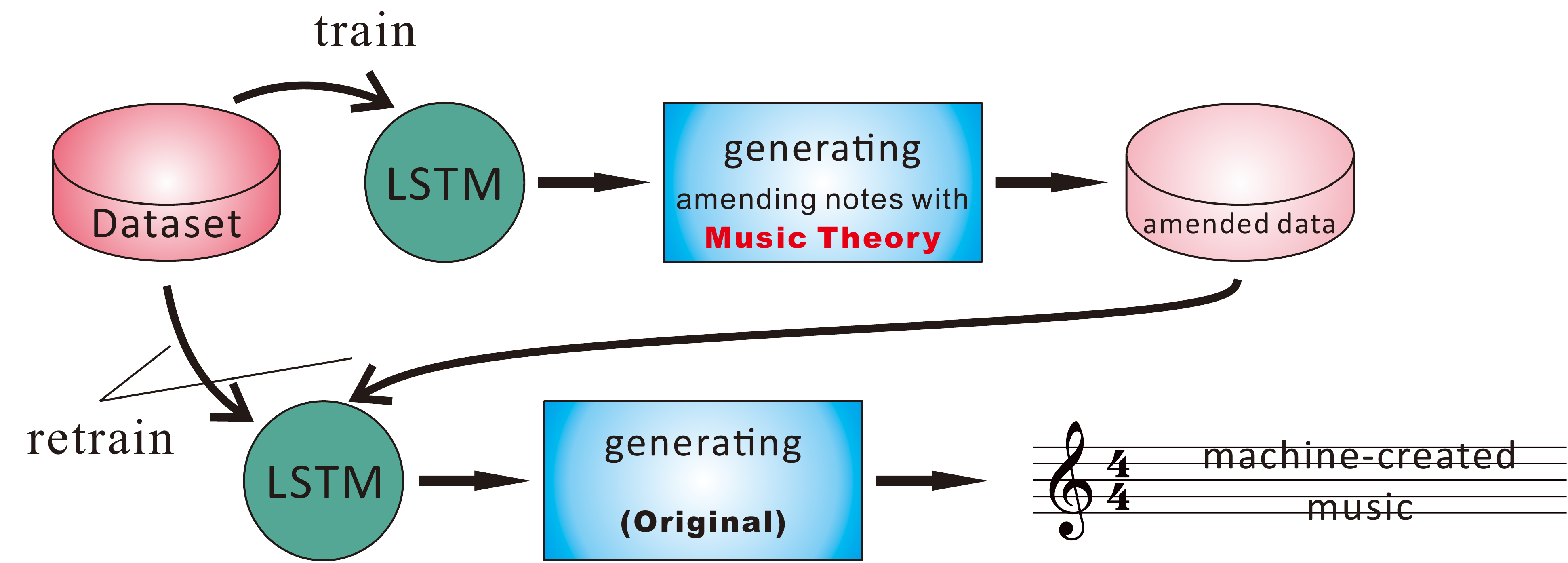}
	\caption{The grammar argumented (GA) method. First, we train the LSTM neural network with the original dataset (top). In the first phase of generation, each note is evaluated with the GA rules from music theory, and each nonconforming note is replaced by a conforming note. The resultant amended data is next mixed with the original dataset, and used to retrain the LSTM network. This network then composes the machine-created music output in the second phase of generation.}
	\label{ga}
	\centering
\end{figure*}

We propose a novel approach called the Grammar Argumented (GA) method that can significantly alleviate this problem without any manual intervention (Fig. \ref{ga}). The idea is to augment the training data such that it also includes machine-generated music that perfectly satisfies the principles of music theory. To do so, the music generation is broken into two phases, the first for generating training data that perfectly conforms to criteria derived from music theory, and the second for the actual musical output. In the first phase, a GA filtering step is applied to the output, such that only melodies satisfying the three grammatical rules described below can pass (as amended data). The residual nonconforming data will be abandoned by resampling. Next the amended data will be added to the training data for retraining the machine before the second phase of generation produces the actual output. 

Inspired by music theory\cite{jane2016}, we put forward three specific rules for the GA filtering. The first rule is that the notes (after translation to C major) must belong to the C major diatonic scale (DIA). Most of western music (and many from other cultures) is based on the diatonic scale consisting of seven distinct tones C, D, E, F, G, A, and B within an octave, among which various harmonies exist. While occassional chromaticity (presence of overtones C\#, D\#, F\#, G\#, and A\#) can add extra color to a musical piece, LSTM generated music without GA argumentation contains too many overtones and consequently sound random and devoid of structure.

The second rule is that the pitch interval between two consecutive notes do not exceed an octave, i.e. that of short pitch interval (SPI). Large jumps in pitch usually sound disruptive and unlyrical, and we leave their artful implementation to future work.

The third rule is that any three consecutive notes must belong to a triad (TRI). Triads are pairs of pitch intervals representing chords, which are of fundamental importance in musical harmony. There are four types of triads, namely the major, minor, augmented and diminished triads, each inducing a different emotional response. Triads are furthermore the building blocks of all seventh chords, which add sophistication to the composition. 

We conclude this subsection by providing a very simple illustration of the GA method. In the first phase of generation, the aim is to generate training music that perfectly conforms to the three abovementioned GA rules. When a nonconforming note is predicted, we return to the output layer of model and resample from the output distribution. This operation is repeated until a GA conforming note is generated. For example, suppose that the last note in the output score is (eighth, A5), and that the newly predicted note is (eighth, B6). The new note B6 violate SPI because the pitch interval spans 14 semitones, which is larger than the octave interval of 12 semitones. Although B6 may have the highest probability in the output layer, we abandon it and resample till we arrive at a GA conforming note, which will then be added to the output score. After this first phase of generation, we mix all amended data with the original training set, as illustrated in Fig. \ref{ga}, and then retrain the model for the second phase of actual generation. The amount of GA-conforming data in the training set determines the extent of chromaticity, lyricalness and harmony in the final output music. We emphasize that the simple implementation of these three music theory rules as GA grammers has been possible owing to note-level encoding.

\section{Experiments}\label{sec:ex}
Our model consists of one LSTM layer and one fully connected layer. The LSTM layer includes 128 cells with input dimension 89, the length of each note's binary representation. There are 89 nodes in the fully connected layer, which is also the output layer. We adopt orthogonal initialization for inner cells and glorot uniform initialization for weights. As suggested by Jozefowicz\cite{jozefowicz2015empirical}, the forget gates bias are initialized with an all-ones matrix. The size of our dataset is 30k, which is divided into batches of 64 to speed up the training process. The loss function is defined with categorical crossentropy, and we use Adam\cite{Kingma2014Adam} to perform gradient descent optimization, with learning rate set to 0.001. We build our model on a high-level neural networks library $Keras$ \cite{chollet2015keras} and use $TensorFlow$\cite{tensorflow2015-whitepaper} as its tensor manipulation library.

This model was first trained with the original dataset with the length of seed phrase set to 7 (notes). The loss stopped decreasing after 400 epochs, and we label the resultant weights as Orig. In the first phase of generation, we used Orig to generate 100k notes for each GA rule and obtained 5759, 5217 and 7931 amended notes respectively. 
Each group of amended data was then mixed with the original dataset to produce three different sets of new training data. A fourth set of new training data was obtained by mixing all these three groups of data (MIX) with the original training data. The model was then retrained with these new data, yielding four new sets of weights labeled DIA, SPI, TRI, and MIX, based on the GA rules they conform to. For statistics analysis, we used a public random seed to generate 100k notes with all five sets of weights, including Orig. Finally, the second phase of generation was performed with these five sets of weights to produce the actual output music.

\section{Results and evaluation} 
\begin{figure*}[!t]
	\centering
	\includegraphics[width=5in]{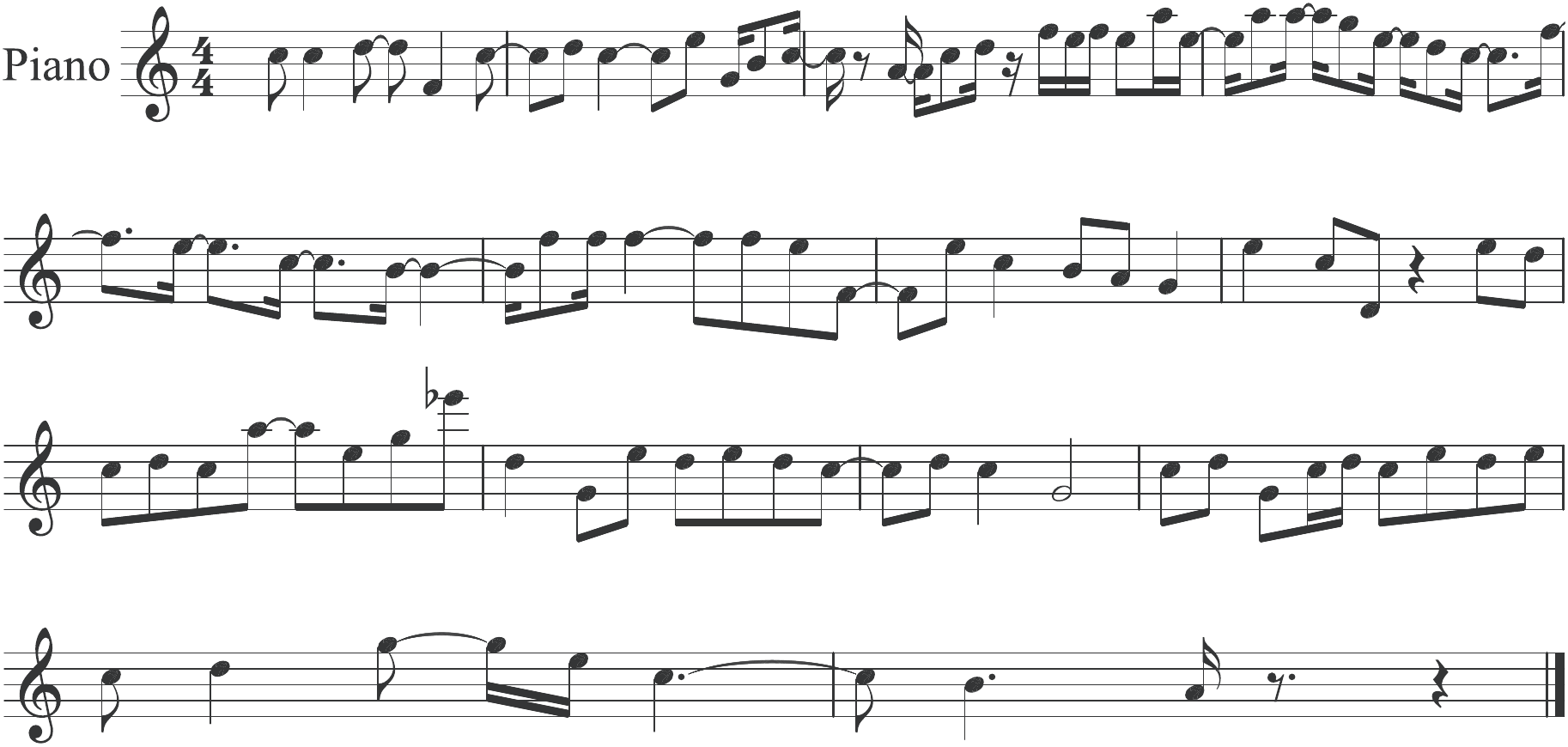}
	\caption{An approximately 100-note segment of the machine's composition. It was generated in MIX mode, which encompasses all three GA rules. 
	}
	\label{result}
	\centering
\end{figure*}

We first look at a representative segment from machine's full composition generated in MIX mode, which encompasses all three GA rules. The music score of this segment is shown in Fig.\ref{result}. Evidently, the machine  prefers notes in the C major diatonic scale, with only one overtone (E-flat). There is also some rudimentary use of repeating motifs, as appearing in bars 3-4 and 10-12. The machine has also employed variations of rhythm in bars 3,4,6 and 12, reminiscent of actual songs. On the whole, the segment is generally lyrical, consistent with music in the dataset.

To quantitatively evaluate the music generated, we put forward three metrics motivated by the GA rules based on music theory (Section \ref{sec:ga}). They are the percentage of notes in the diatonic scale ($p_{dia}$), percentage of pitch intervals within one octave ($p_{SPI}$) and percentage of triads ($p_{tri}$). These metrics are generically applicable for all types of music, and not just those defined by note-level encoding.

\subsection{$p_{dia}$}
\begin{table}[!t]
	\renewcommand{\arraystretch}{1.3}
	\caption{$p_{dia}$ (\%) of Dataset (DS) and outputs from the Five Modes}
	\label{pdia}
	\centering
	\begin{tabular}{c|cccccc}
		\hline
		& DS & Orig & DIA & SPI & TRI & MIX \\
		\hline
		C & 8.9 & 6.6 & 11.7 & 8.6 & 6.2 & 10.8 \\
		D & 7.8 & 6.4 & 12.1 & 7.9 & 4.9 & 9.4 \\
		E & 9.1 & 7.5 & 14.5 & 9.2 & 7.8 & 11.7 \\
		F & 7.6 & 7.3 & 8.2 & 7.9 & 7.4 & 7.4 \\
		G & 7.0 & 5.2 & 10.0 & 7.3 & 5.0 & 8.3 \\
		A & 6.6 & 5.4 & 9.8 & 7.1 & 4.7 & 8.5 \\
		B & 8.0 & 7.5 & 8.6 & 8.1 & 6.9 & 7.9 \\
		\hline
		Total ($p_{dia}$) & 54.8 & 45.9 & \textbf{75.0} & 56.1 & 42.9 & \textbf{64.0} \\
		\hline
	\end{tabular}
\end{table}

Our results show that music generated in the DIA mode indeed possess significantly more notes adhering to the diatonic scale. From Table \ref{pdia}, which displays the percentages of each of the seven tones in C major diatonic scale, $p_{dia}$ is 29.1 percentage points higher in the DIA mode than in Orig, where the DIA grammer rules have not been applied. Indeed the DIA GA method can significantly decrease the occurence of overtones, even if the original dataset contain key changes and depart significantly from the original diatonic scale (as seen from its relatively low $p_{dia}$). Incidentally, the tonic note C is observed to have one of the highest occurrences, in line with expectations from more advanced music theory beyond the GA rules. 

\subsection{$p_{SPI}$}
\begin{table}[!t]
	\renewcommand{\arraystretch}{1.3}
	\caption{$1-p_{SPI}$ (\%) of Dataset (DS) and outputs from the Five Modes}
	\label{pspi}
	\centering
	\begin{tabular}{c|c|c|c|c|c|c}
		\hline
		& DS & Orig & DIA & SPI & TRI & MIX \\
		\hline
		$p_{SPI}$ & 12.9 & 14.2 & 12.3 & \textbf{9.4} & 13.2 & \textbf{10.2} \\
		\hline
	\end{tabular}
\end{table}

In Table \ref{pspi}, we tabulate the percentage of pitch intervals within an octave for all the various mode outputs. A high $p_{SPI}$ percentage corresponds to a more lyrical composition. Evidently, the SPI and MIX modes produces music with the highest $p_{SPI}$, such that there are about 30 percent fewer pitch jumps larger than one octave than that of Orig mode, whose data was generated before any GA rule has been applied.

\subsection{$p_{tri}$}
\begin{table}[!t]
	\renewcommand{\arraystretch}{1.3}
	\caption{$p_{tri}$ (\%) of Dataset (DS) and outputs from the Five Modes}
	\label{ptri}
	\centering
	\begin{tabular}{c|cccccc}
		\hline
		& DS & Orig & DIA & SPI & TRI & MIX \\
		\hline
		Major & 2.3 & 2.2 & 2.3 & 2.4 & 7.9 & 5.8 \\
		Minor & 2.1 & 2.0 & 2.3 & 2.1 & 7.7 & 5.6 \\
		Augmented & 0.0 & 0.1 & 0.1 & 0.2 & 0.5 & 0.4 \\
		Diminished & 0.2 & 0.3 & 0.4 & 0.5 & 1.3 & 1.1 \\
		\hline
		Total ($p_{tri}$) & 4.6 & 4.6 & 5.1 & 5.3 & \textbf{17.4} & \textbf{12.9} \\
		\hline
	\end{tabular}
\end{table}

Our results in Table \ref{ptri} show that the music composed in the TRI and MIX modes indeed contain more triads than the other modes'. $p_{tri}$, the percentage of triads, is computed by counting the total proportion of 3 consecutive notes assuming one of the four types of triads. The TRI mode, in particular, generates music within an almost fourfold increase in the number of triads.

Note that the music composed in MIX mode perform well under all three metrics. This suggests that the three GA rules are not conflicting, but rather are complementary ingredients for lyrical music. We emphasize that although some of the training data satisfy these metrics perfectly by construction, the final output music is generated purely by machine learning, and without human intervention.

\section{Conclusion}
By themselves, simple LSTM neural networks cannot generate music that are appealing from the standpoint of music theory. We addressed this problem by augmenting the training data with grammar argumented (GA) machine generated ouput. In this way, the machine can be trained to generate music that inherits the naturalness of the original dataset while closely adhering to the major aspects of music theory. Since the GA filters are applied to the training data and not directly to the output, the latter is still generated by a completely bona-fide machine learning approach. Our note-level encoding method also allows a more authentic emulation of human composers, as well as provide a natural platform for implementing our grammar argumented method. The generated music generally sound lyrical, and adhere well to music theory according to the three major criteria we proposed.

\section*{Acknowledgment}
The authors will like to thank Chuangjie Ren and Qingpei Liu for helpful discussions on neural networks.

\ifCLASSOPTIONcaptionsoff
  \newpage
\fi

\bibliographystyle{IEEEtran}
\bibliography{mycitation}


\end{document}